\documentclass[pdflatex,sn-mathphys-num]{sn-jnl}

\usepackage{graphicx}%
\usepackage{multirow}%
\usepackage{amsmath,amssymb,amsfonts}%
\usepackage{amsthm}%
\usepackage{mathrsfs}%
\usepackage[title]{appendix}%
\usepackage{xcolor}%
\usepackage{textcomp}%
\usepackage{manyfoot}%
\usepackage{booktabs}%
\usepackage{algorithm}%
\usepackage{algorithmicx}%
\usepackage{float}
\usepackage{algpseudocode}%
\usepackage{listings}%

\begin{document}

\title[Article Title]{Social Determinants of Health and Fentanyl Overdose Mortality Across US Counties: An XGBoost and SHAP Analysis Identifying Silent Risk Counties and Treatment Deserts}

\author*[1]{\fnm{Kabi Raj} \sur{Tiruwa}}\email{kabirajtiruwa5@gmail.com}

\author[1]{\fnm{Abhisan} \sur{Ghimire}}\email{abeesun@gmail.com}
\equalcont{These authors contributed equally to this work.}

\author[2]{\fnm{Anuj Kumar} \sur{Shah}}\email{anuj.shah260@gmail.com}
\equalcont{These authors contributed equally to this work.}

\affil*[1]{\orgdiv{Clark University}, \orgaddress{\street{910 Main St}, \city{Worcester}, \postcode{01610}, \state{MA}, \country{USA}}}

\affil[2]{\orgdiv{Yeshiva University}, \orgaddress{\street{500 W 185th St}, \city{New York}, \postcode{10033}, \state{NY}, \country{USA}}}

\abstract{
\textbf{Background}: Fentanyl overdose deaths are still increasing across the U.S. Even though the 
crisis is growing, we still do not fully understand which county-level social and 
structural conditions lead to higher overdose death rates. Social determinants of 
health, including disability, treatment access, and behavioral health issues, may 
help identify vulnerable counties before deaths become severe. No earlier study 
has used explainable machine learning with SHAP attribution on 2022 CDC WONDER 
data to study treatment access gaps and silent risk counties.

\par\medskip
\textbf{Methods}: We combined data from four government sources for 975 U.S. counties where 
overdose data was available, including CDC WONDER (2022) overdose mortality data, 
CDC Social Vulnerability Index (SVI), CDC PLACES health behavior data, and Area 
Health Resources Files for healthcare workforce and resources. An XGBoost machine 
learning model was used to predict overdose mortality risk using Standardized 
Mortality Ratio (SMR) for each county. Five-fold cross-validation was used to 
test model accuracy, and SHAP values were used to show which factors increase or 
decrease risk.

\par\medskip
\textbf{Results}: The XGBoost model performed better than all other tested models in predicting overdose risk. It matched real rankings fairly well (Spearman $\rho$ = 0.67), explained about 46\% of the variation in deaths (R$^{2}$ = 0.457), had moderate prediction error (MAE = 0.409), and correctly identified about 71\% of high-risk counties. The most important predictors were disability rate, hypertension, smoking and lack of vehicle access. Treatment desert counties 
had much higher overdose mortality (52.6\% higher; SMR 1.786 vs 1.170; 
$p<0.0001$). K-means found 143 silent risk counties with high risk but not yet 
high deaths. Overdose deaths were geographically clustered (Moran's I = 0.5053, 
$p$=0.001) with 75 hotspots and 136 coldspots. Suppressed counties were 58.2\% 
of 2022 CDC WONDER counties, mostly rural (72\%) and treatment deserts (65\%).

\par\medskip
\textbf{Conclusions}: County-level SDOH factors can help predict overdose deaths, especially disability 
level, access to treatment, and behavioral health burden. These factors identify 
counties at risk before the crisis becomes severe. MOUD expansion should focus on 
treatment desert counties, and silent risk counties should be monitored closely 
for early prevention.
}


\maketitle

\section{Introduction}\label{sec1}
\setlength{\parskip}{1em}
The drug crisis has become one of the major issues in recent years in the United States. Over the past two decades, the number of deaths caused by drug overdose has increased almost five times \cite{bib1}. A large part of these deaths is related to fentanyl (synthetic opioids), which has shown a very sharp rise from 2013 to 2021 \cite{bib2}. The epidemic in the U.S. started first as a prescription drug problem and later changed into a fentanyl crisis. It can be understood in three main phases. The first wave began in the late 1990s with prescription opioids. The second wave started around 2010 with heroin use. The third wave began in 2013 with synthetic opioids like fentanyl, which is the most dangerous. Now, with the increase in deaths related to stimulants, it seems like we may be entering a fourth wave \cite{bib3,bib4}. When we look at this issue more closely, we can see that drug overdose deaths are not the same everywhere. There are big differences at the county level across the country \cite{bib5}.

Social Determinants of Health (SDoH) have been shown to act as important predictors of opioid overdose deaths. Empirical findings suggest a clear relationship between county-level SDoH factors and mortality from opioid use, and identifying which SDoH factors drive these deaths is an important step for targeted county-level interventions \cite{bib5,bib6}. Counties with factors such as higher income inequality, elevated crime rates, and lower access to internet and healthcare tend to have higher overdose death rates \cite{bib7}. From 2019 to 2020, drug overdose deaths increased sharply, rising by 44\% among non-Hispanic Black individuals and 39\% among non-Hispanic American Indian or Alaska Native individuals. Clear differences were seen across age, sex, and race. In 2020, Black males aged 65 and older had nearly seven times higher overdose death rates compared to White males in the same age group. Most cases involved a history of substance use, but Black individuals had the lowest rate of prior treatment (8.3\%). These disparities were especially severe in counties with higher income inequality \cite{bib8}. Considering all this evidence, it is important to identify which specific SDoH factors are most important and which counties are structurally at higher risk for crisis. This requires methods that go beyond simple regression models.

Counties have been classified as high-risk when they have both high opioid overdose mortality and low OUD (opioid use disorder) treatment capacity. One study used a geospatial cross-sectional analysis with county-level data from 2015 to 2017 to identify such counties \cite{bib9}. However, the extent to which lack of treatment alone drives mortality, independent of SDoH burden, has not been studied using explainable machine learning methods. Another study used regression analysis with a county-level predictive approach based on restricted CDC mortality data, and the model was validated \cite{bib10}. However, this approach mainly relied on geographic patterns rather than SDoH features as the main predictors. Recently, XGBoost with SHAP attribution has been applied for county-level overdose prediction \cite{bib11}, but that study used multi-year data and did not clearly address the issue of systematically suppressed county-level population data.

This study addresses these gaps by using publicly available 2022 data combined from four federal sources. Specifically, we aim to identify county-level SDoH predictors of fentanyl overdose mortality using XGBoost with SHAP-based feature attribution. We also calculate the mortality impact in treatment desert counties and identify counties at ``silent risk,'' meaning those with high structural burden but below-median current mortality. In addition, we examine the geographic clustering of overdose mortality and study the demographic profile of counties with suppressed data, which are often systematically excluded.

\section{Method}\label{sec2}

\subsection{Data Source}\label{subsec2a}

For this study, we used publicly available 2022 data from four federal sources. The CDC WONDER Multiple Cause of Death database provided county-level mortality rates using UCD codes X40--X44, Y10--Y14, and MCD code T40.4. Social Vulnerability data from the CDC/ATSDR included 16 indicators across four SDoH domains at the county level. The CDC PLACES dataset provided modeled estimates of health behaviors and chronic conditions, mainly based on BRFSS data. The HRSA Area Health Resources Files (AHRF) 2023--2024 provided county-level data on healthcare workforce and infrastructure, from which four variables were extracted. All four datasets were merged at the county level using FIPS (Federal Information Processing Standard) codes.

\subsection{Data Preparation}\label{subsec2b}

In the CDC WONDER mortality data, counties with fewer than 10 overdose deaths are marked as suppressed. Out of 2,351 WONDER counties, 1,368 were suppressed and excluded from the primary predictive analysis. This study analyzes 975 non-suppressed counties. Counties with zero psychiatrists, based on AHRF data, were classified as treatment desert counties. The absence of a MOUD treatment provider acts as a structural barrier to accessing treatment \cite{bib9}.

\subsection{Feature Engineering}\label{subsec2c}

The outcome variable in this study was the 2022 standardized mortality ratio (SMR), 
which was extracted from CDC WONDER. The SMR was calculated as:

\begin{equation}
\text{SMR} = \frac{\text{Observed Opioid Overdose Deaths}}{\text{Expected Deaths}}
\end{equation}

where expected deaths were derived as:

\begin{equation}
\text{Expected Deaths} = \text{County Population} \times 
\frac{\text{National Reference Rate}}{100{,}000}
\end{equation}

The national reference rate (23.91 per 100,000) was calculated from the total observed 
deaths and population across all 975 non-suppressed counties in the 2022 CDC WONDER 
dataset. An SMR of 1.0 indicates a county's mortality equals the national average; 
values above 1.0 indicate excess mortality and values below 1.0 indicate below-average 
mortality. The observed SMR ranged from 0.11 to 5.86 across the analytic sample.

In total, 25 county-level predictors were collected from four federal data sources and grouped into four main domains. The first domain, socioeconomic deprivation, included variables such as poverty rate, unemployment, housing burden, lack of high school education, uninsured rate, and 
the CDC SVI composite percentile. The second domain, social vulnerability, included age 65 and above, disability, single-parent households, minority population, mobile home residency, and lack of vehicle access. The third domain covered health behaviors and 
chronic conditions, including depression, smoking, obesity, poor mental health, poor sleep, physical inactivity, binge drinking, diabetes, and hypertension prevalence from CDC PLACES. The fourth domain focused on treatment infrastructure, including psychiatrist density, primary care density, treatment desert status, and rural classification from AHRF. Missing values across all predictors were handled by imputing with column-wise medians using scikit-learn.

\subsection{Machine Learning and SHAP Analysis}\label{subsec2d}

In this study, XGBoost \cite{bib12} was selected because it can handle complex nonlinear relationships between variables and also performs well when predictors are correlated. This model has also shown strong performance in county-level opioid mortality prediction \cite{bib11}. The model performance was tested using five-fold cross-validation, and it was evaluated using six metrics: R$^{2}$, RMSE, MAE, MAPE, Spearman correlation, and high-risk recall. A total of four models were compared under the same conditions, including Linear Regression, LASSO and Random Forest. Among these, XGBoost achieved the best predictive performance and was selected for further SHAP analysis. For model interpretation, SHAP analysis was used to understand how each feature affected the predictions. SHAP TreeExplainer was applied to generate beeswarm plots for the most important features and dependence plots for the top four predictors. In addition, spatial autocorrelation was computed using the \texttt{esda.Moran\_Local} function from the PySAL ecosystem.

\subsection{Clustering and Spatial Autocorrelation}\label{subsec2e}

K-Means clustering was used to group counties based on similar structural and social conditions. The best value for K was selected using the silhouette coefficient by testing K=2 to K=8, which measured how well counties fit into their groups. After clustering, silent risk counties were identified as counties with high structural burden but below-median overdose mortality. This means these counties may not currently have high overdose deaths, but their conditions make them vulnerable to future crisis. This created a four-quadrant framework: high burden/high mortality (active crisis), high burden/low mortality (silent risk), low burden/high mortality (unusual high mortality), and low burden/low mortality (lower risk). Moran's I was used to measure whether overdose deaths were geographically clustered. Higher Moran's I values showed that nearby counties had similar overdose rates, while values near zero suggested random patterns. Queen contiguity weights defined neighboring counties as those sharing a border or corner. LISA (Local Indicators of Spatial Association) was used to identify specific local clusters, including hotspots (high-overdose counties surrounded by high-overdose counties) and coldspots (low-overdose counties surrounded by low-overdose counties). Statistical significance was tested at $p < 0.05$ using permutation testing to confirm that these patterns were not due to chance. Because this study used only publicly available, de-identified county-level data, no patient consent or IRB approval was required.

\section{Results}\label{sec3}

\subsection{Sample Characteristics}\label{subsec3a}

This study included 975 U.S. counties with non-suppressed 2022 CDC WONDER overdose mortality data, representing 31.0\% of all 3,143 U.S. counties. Many smaller or low-population counties were excluded because the CDC suppresses data when death counts are very low to protect privacy. County characteristics stratified by K-means cluster assignment (K=2, silhouette = 0.214) are presented in Table~\ref{tab:characteristics}. Cluster 1 (n=438 counties) showed significantly higher mean overdose SMR (1.557 vs 0.999), greater socioeconomic deprivation, higher behavioral health burden, and lower healthcare access compared to Cluster 0 (n=537 counties). Significant differences were found across 22 of 24 SDoH indicators ($p<0.05$ for 22 of 24 indicators; Table~\ref{tab:characteristics}). Housing burden and age 65+ were the only factors that did not differ significantly between the clusters (p=0.12 and p=0.27, respectively).

\begin{table}[h]
\caption{County Characteristics Stratified by K-Means Cluster Assignment}\label{tab:characteristics}
\begin{tabular}{@{}llllll@{}}
\toprule
Variable & Full Sample & Cluster 0 & Cluster 1 & p-value & Sig \\
\midrule
SMR (obs/exp) & $1.25 \pm 0.79$ & $1.00 \pm 0.51$ & $1.56 \pm 0.95$ & 0.0000 & *** \\
Poverty Rate (\%) & $21.90 \pm 7.14$ & $17.39 \pm 4.66$ & $27.42 \pm 5.62$ & 0.0000 & *** \\
Unemployment Rate (\%) & $5.27 \pm 1.61$ & $4.63 \pm 1.17$ & $6.06 \pm 1.72$ & 0.0000 & *** \\
Housing Burden (\%) & $24.44 \pm 4.69$ & $24.10 \pm 4.19$ & $24.85 \pm 5.22$ & 0.1195 & ns \\
No HS Diploma (\%) & $10.49 \pm 4.44$ & $8.08 \pm 2.77$ & $13.45 \pm 4.32$ & 0.0000 & *** \\
Uninsured Rate (\%) & $8.16 \pm 3.68$ & $6.74 \pm 2.79$ & $9.90 \pm 3.87$ & 0.0000 & *** \\
SVI Percentile & $0.54 \pm 0.27$ & $0.39 \pm 0.22$ & $0.71 \pm 0.21$ & 0.0000 & *** \\
Age 65+ (\%) & $17.98 \pm 4.33$ & $18.02 \pm 4.70$ & $17.94 \pm 3.83$ & 0.2658 & ns \\
Disability Rate (\%) & $14.86 \pm 3.83$ & $12.96 \pm 2.76$ & $17.19 \pm 3.67$ & 0.0000 & *** \\
Single-Parent HH (\%) & $6.21 \pm 1.83$ & $5.41 \pm 1.18$ & $7.19 \pm 2.00$ & 0.0000 & *** \\
Minority Pop (\%) & $29.15 \pm 19.65$ & $26.42 \pm 16.39$ & $32.49 \pm 22.60$ & 0.0019 & ** \\
Mobile Homes (\%) & $9.12 \pm 8.22$ & $5.29 \pm 4.43$ & $13.81 \pm 9.31$ & 0.0000 & *** \\
No Vehicle Access (\%) & $6.57 \pm 4.67$ & $6.21 \pm 5.23$ & $7.01 \pm 3.85$ & 0.0000 & *** \\
Depression Prev (\%) & $22.03 \pm 3.29$ & $20.85 \pm 2.94$ & $23.48 \pm 3.12$ & 0.0000 & *** \\
Current Smoking (\%) & $18.64 \pm 4.16$ & $16.24 \pm 3.09$ & $21.58 \pm 3.34$ & 0.0000 & *** \\
Obesity Rate (\%) & $34.61 \pm 4.92$ & $31.85 \pm 4.30$ & $37.99 \pm 3.23$ & 0.0000 & *** \\
Poor Mental Health (\%) & $15.69 \pm 1.85$ & $14.64 \pm 1.39$ & $16.97 \pm 1.50$ & 0.0000 & *** \\
Poor Sleep (\%) & $34.78 \pm 3.43$ & $32.91 \pm 2.74$ & $37.09 \pm 2.73$ & 0.0000 & *** \\
Physical Inactivity (\%) & $24.29 \pm 4.68$ & $21.18 \pm 3.08$ & $28.11 \pm 3.26$ & 0.0000 & *** \\
Binge Drinking (\%) & $17.26 \pm 2.52$ & $18.14 \pm 2.59$ & $16.19 \pm 1.95$ & 0.0000 & *** \\
Diabetes Prev (\%) & $10.16 \pm 1.86$ & $8.94 \pm 1.07$ & $11.65 \pm 1.50$ & 0.0000 & *** \\
High Blood Pressure (\%) & $32.00 \pm 4.44$ & $29.21 \pm 2.81$ & $35.34 \pm 3.64$ & 0.0000 & *** \\
Psychiatrists /100k & $8.33 \pm 10.95$ & $10.86 \pm 12.66$ & $5.23 \pm 7.27$ & 0.0000 & *** \\
Primary Care /100k & $58.28 \pm 30.87$ & $67.24 \pm 31.89$ & $47.28 \pm 25.62$ & 0.0000 & *** \\
Treatment Desert, n (\%) & 127 (13.0\%) & 26 (4.8\%) & 101 (23.1\%) & 0.0000 & *** \\
Rural County, n (\%) & 260 (26.7\%) & 83 (15.5\%) & 177 (40.4\%) & 0.0000 & *** \\
\bottomrule
\end{tabular}
\end{table}
\vspace{-2.5em}
\noindent\footnotesize\textit{Note:} SMR: Standardized Mortality Ratio; SVI: Social Vulnerability Index. Significance: *** $p < 0.001$, ** $p < 0.01$, * $p < 0.05$, ns = not significant.
\normalsize

\subsection{Model Performance}\label{subsec3b}

XGBoost achieved the highest predictive performance among all four evaluated models (Table~\ref{tab:performance}). The model showed strong geographic risk stratification, correctly identifying 71.1\% of counties with above-average mortality (SMR $> 1.0$) within the upper half of predicted risk (high-risk recall). The Spearman rank correlation between predicted and observed SMR was $\rho$=0.670 ($p<0.001$), with a Pearson correlation of r=0.687. These results suggest that if the model were used as a geographic triage tool to prioritize intervention resources, it would correctly direct resources to about 7 out of 10 truly high-burden counties. For point prediction, the model achieved R$^{2}$ = 0.457 ($\pm$0.049), RMSE = 0.574 ($\pm$0.048 SMR units), and MAE = 0.409 SMR units. The mean absolute percentage error was 42.6\%, and 43.3\% of county predictions were within acceptable prediction range, suggesting moderate prediction error but useful geographic targeting ability. XGBoost outperformed all three comparison models across all metrics, including Random Forest (R$^{2}$=0.421, RMSE=0.592), Linear Regression (R$^{2}$=0.376, RMSE=0.613), and LASSO (R$^{2}$=0.372, RMSE=0.616). The very similar performance between LASSO and Linear Regression suggests that simple linear models were not sufficient for this problem, supporting the use of gradient-boosted tree methods (Table~\ref{tab:performance}).

\begin{table}[h]
\caption{Comparison of Model Performance for Predicting County-Level SMR}\label{tab:performance}
\begin{tabular}{@{}llll@{}}
\toprule
Model & $R^{2} \pm \text{SD}$ & $\text{RMSE} \pm \text{SD}$ & MAE (SMR) \\
\midrule
XGBoost & $0.457 \pm 0.049$ & $0.5737 \pm 0.0482$ & 0.4087 \\
Random Forest & $0.421 \pm 0.059$ & $0.5919 \pm 0.0470$ & 0.4308 \\
Linear Regression & $0.376 \pm 0.089$ & $0.6133 \pm 0.0534$ & 0.4441 \\
LASSO & $0.372 \pm 0.080$ & $0.6163 \pm 0.0564$ & 0.4412 \\
\bottomrule
\end{tabular}
\end{table}
\vspace{-2.5em}
\noindent\footnotesize\textit{Note:} $R^{2}$: Coefficient of Determination; RMSE: Root Mean Square Error; MAE: Mean Absolute Error; SMR: Standardized Mortality Ratio; SD: Standard Deviation.
\normalsize

\subsection{SHAP Feature Attribution}\label{subsec3c}

\begin{figure}[H]
\centering
\includegraphics[width=\textwidth]{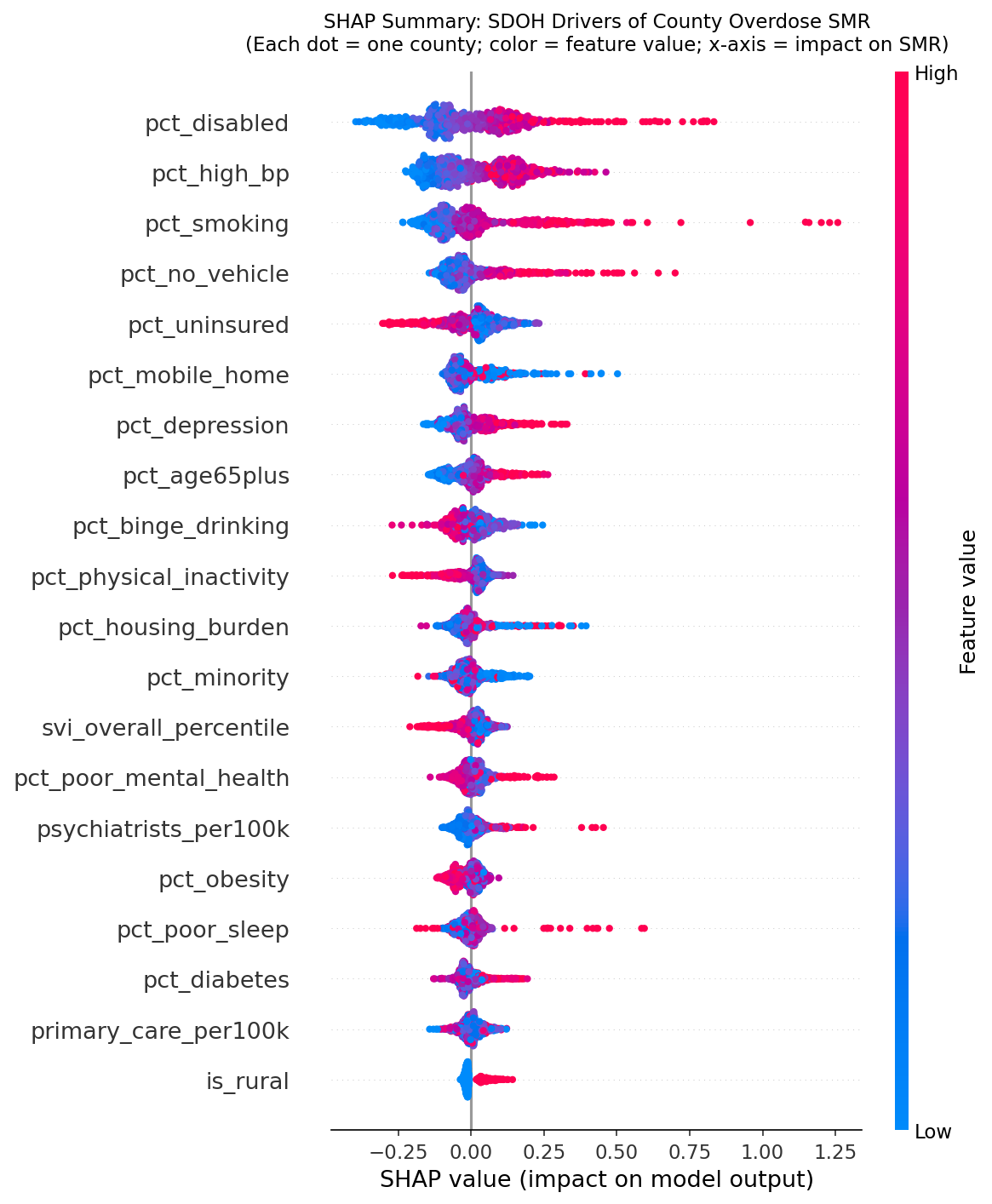}
\caption{SHAP summary beeswarm plot showing the global influence of the top predictors on county-level fentanyl overdose mortality. Features are ranked by mean absolute SHAP value.}\label{fig:beeswarm}
\end{figure}

\begin{figure}[H]
\centering
\includegraphics[width=0.8\textwidth]{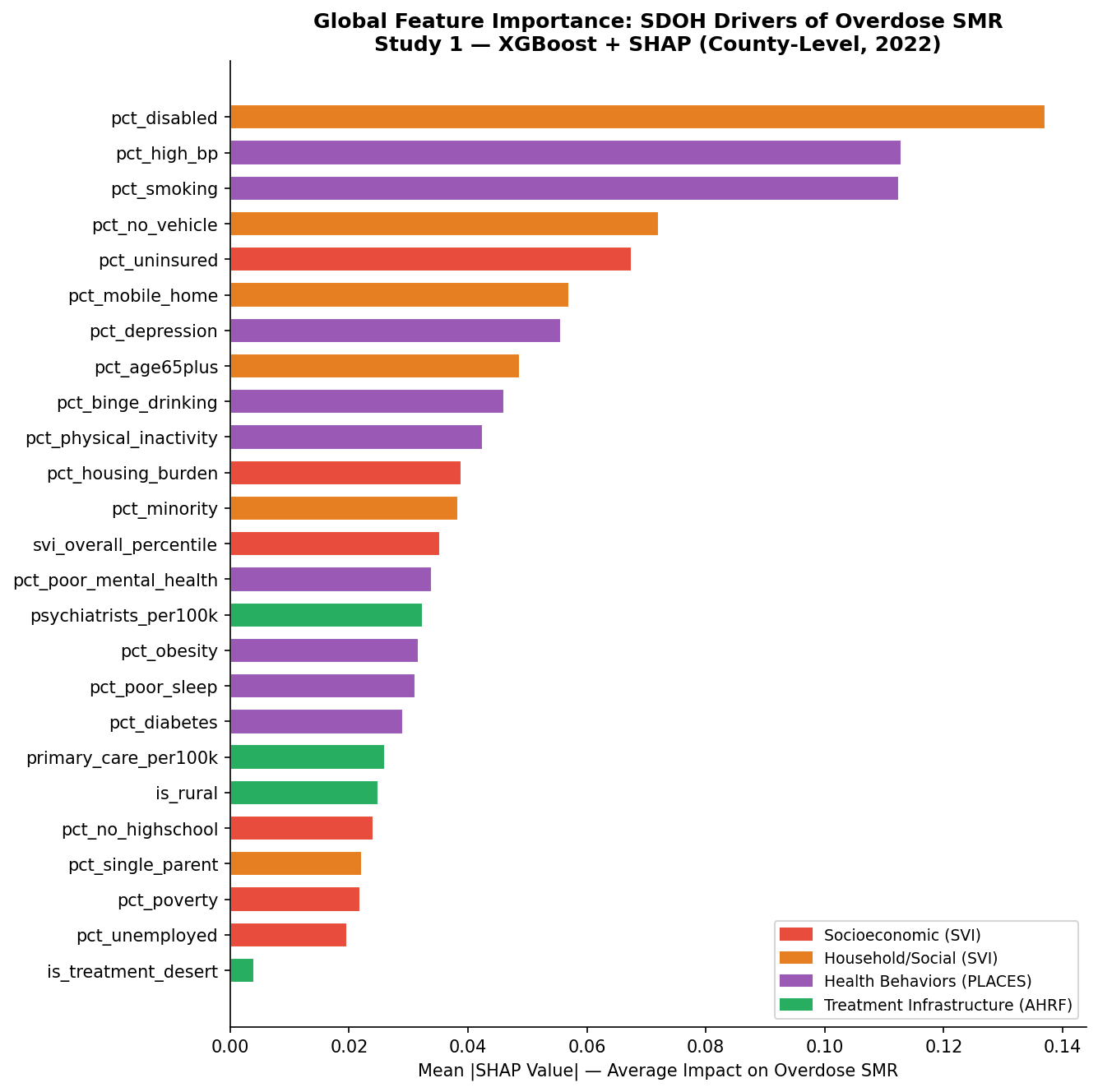}
\caption{Global feature importance ranking. The x-axis represents the average impact (mean absolute SHAP value) of each social determinant of health on the model's output.}\label{fig:importance}
\end{figure}

County-level factors were identified as important predictors of overdose mortality using XGBoost with SHAP analysis. SHAP TreeExplainer showed that disability rate had the strongest influence on predicting county overdose mortality (Figure~\ref{fig:beeswarm}). The four leading predictors, ranked by mean absolute SHAP value, were: (1) disability rate, (2) hypertension prevalence, (3) smoking rate, and (4) percentage of households without vehicle access (Figure~\ref{fig:importance}).

\begin{figure}[H]
\centering
\includegraphics[width=\textwidth]{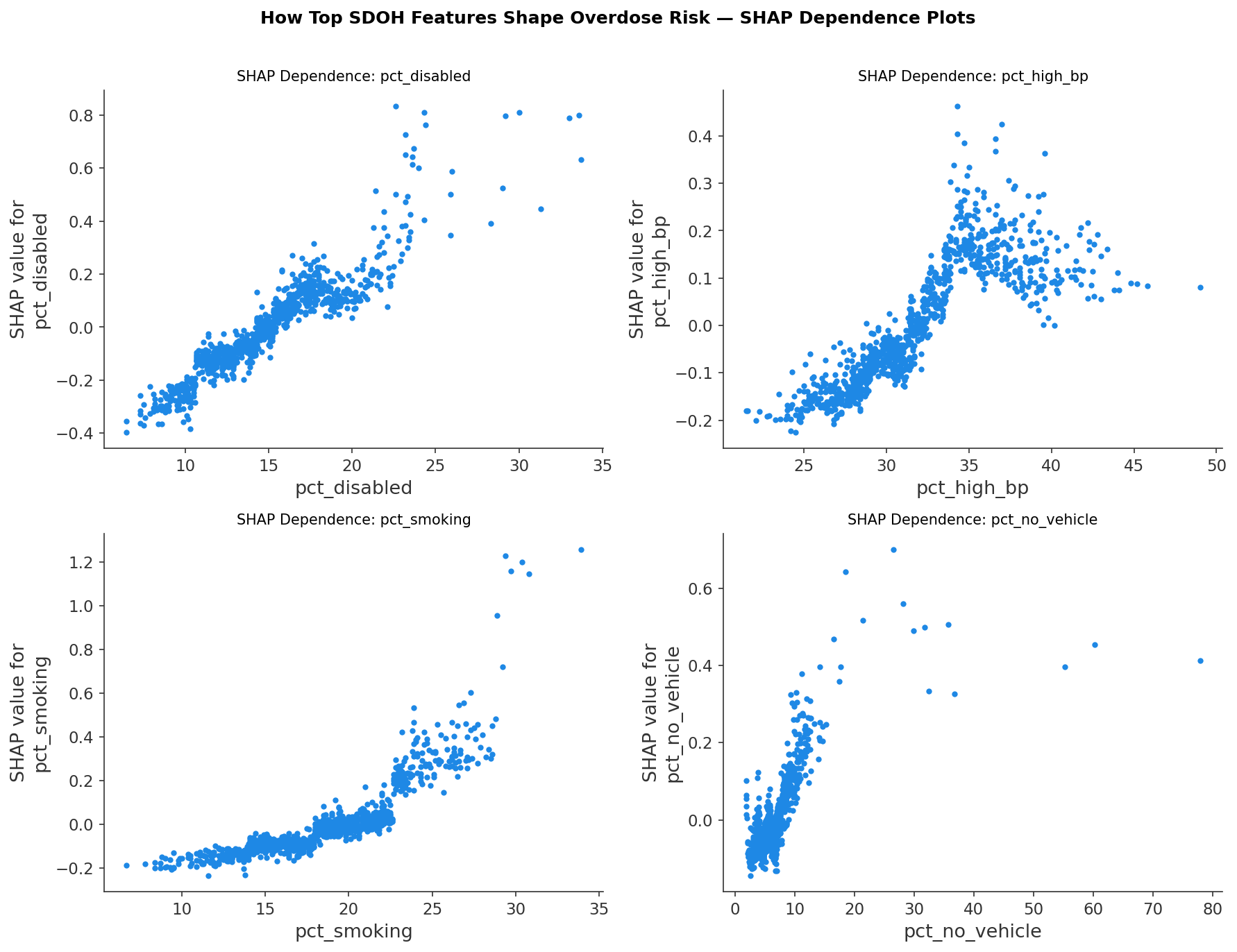}
\caption{SHAP dependence plots showing the nonlinear relationships between mortality and the top four predictors: (A) disability rate, (B) high blood pressure, (C) smoking prevalence, and (D) no vehicle access.}\label{fig:dependence}
\end{figure}

All of these predictors showed a positive relationship with overdose mortality, meaning that higher levels of these factors were associated with higher predicted overdose death rates. SHAP dependence plots (Figure~\ref{fig:dependence}) also showed nonlinear threshold effects, especially for disability rate and smoking rate. This means the risk did not increase in a simple linear pattern, but instead rose more sharply after certain levels. These findings suggest that counties with very high disability or smoking rates may experience a much greater overdose burden than expected.

\subsection{Treatment Desert Penalty}\label{subsec3d}

\begin{figure}[H]
\centering
\includegraphics[width=\textwidth]{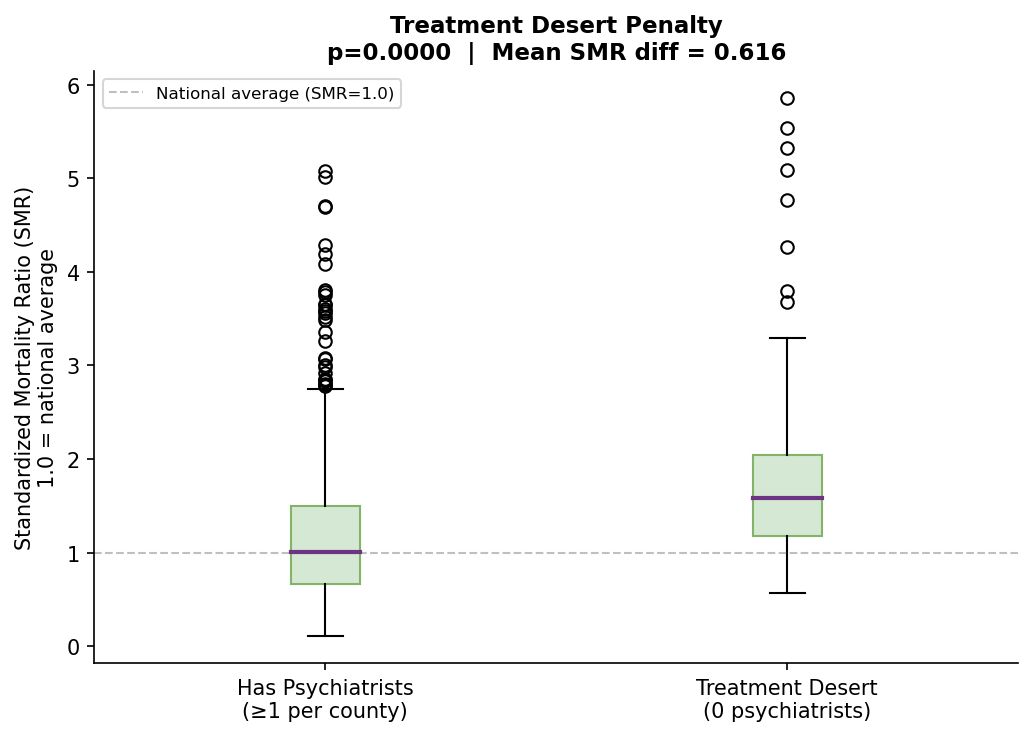}
\caption{Comparison of mean overdose mortality between treatment desert counties (zero psychiatrists) and non-desert counties. Treatment desert counties showed significantly higher SMR than non-desert counties ($p<0.0001$).}\label{fig:treatment}
\end{figure}

Counties designated as treatment deserts, defined as having zero psychiatrists per AHRF (n=127 counties, 13.0\% of the analytic sample), showed significantly higher mean overdose mortality than non-desert counties (SMR 1.786 vs. 1.170; difference: +0.616 SMR; $p<0.0001$, independent sample t-test; Figure~\ref{fig:treatment}). These counties had very limited or no psychiatrist access, which may reduce access to behavioral health and substance use treatment and contribute to higher overdose deaths.

This 52.6\% mortality penalty means that treatment desert counties had about 52\% higher overdose mortality compared to non-desert counties. Among treatment desert counties, 79.5\% (n=101) were also classified as high-burden counties in the K-Means analysis, showing that many of these counties faced both high social vulnerability and lack of treatment access.

These findings suggest that lack of treatment infrastructure may increase overdose mortality beyond other social risk factors.

\subsection{Silent Risk Counties}\label{subsec3e}

Counties were grouped into four risk categories based on two main measures: the Social Determinants of Health (SDoH) burden score and the current overdose mortality rate (SMR). The SDoH burden score reflects structural and social disadvantages such as poverty, disability, limited access to healthcare, smoking rates, transportation barriers, and the overall Social Vulnerability Index (SVI). The mortality threshold was set at the median SMR (1.070), and the burden threshold was set at the 60th percentile of the composite score.

\begin{figure}[H]
\centering
\includegraphics[width=\textwidth]{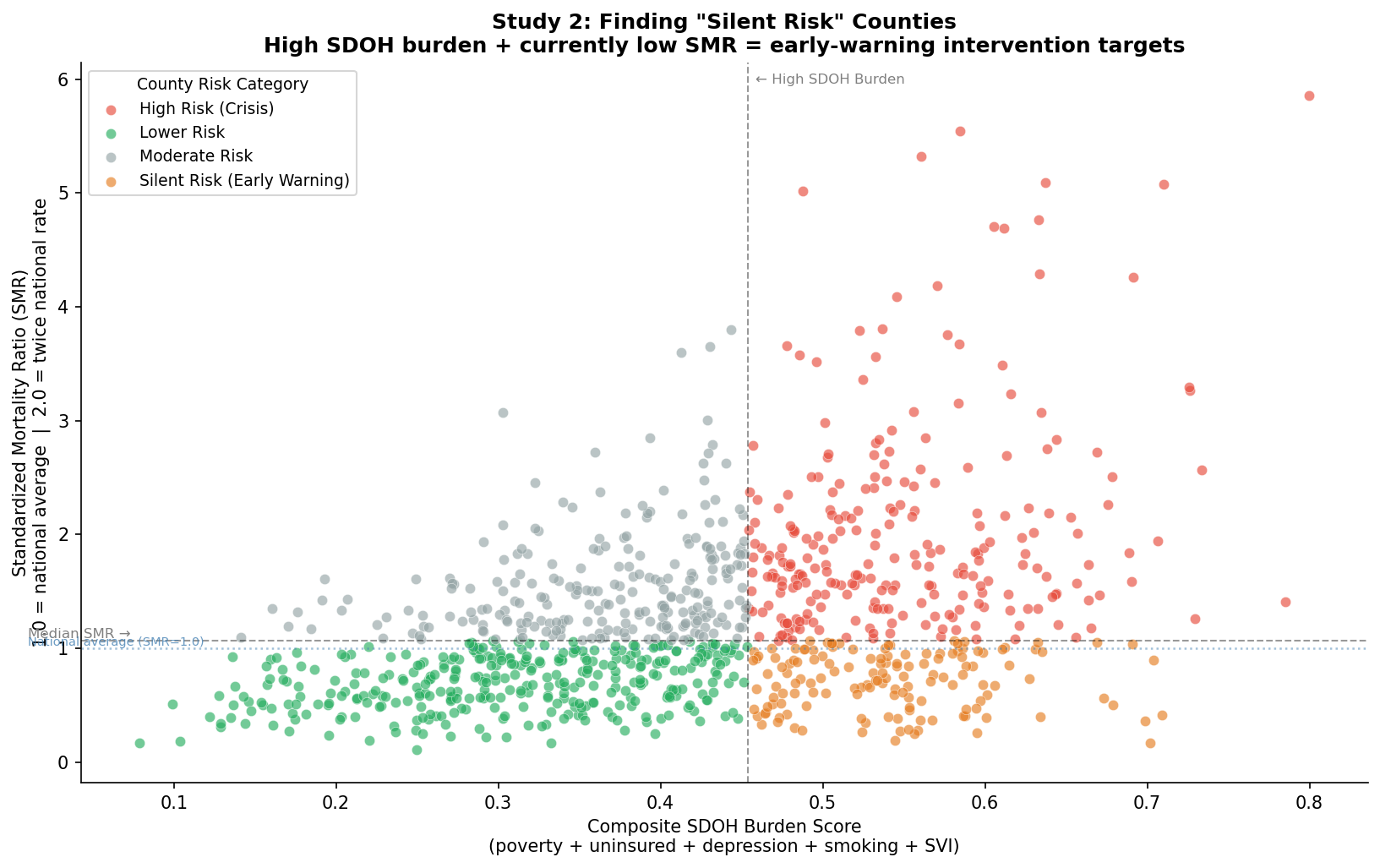}
\caption{Four-quadrant framework of county-level overdose risk. Counties are classified into High Risk/Crisis (n=247), Silent Risk/Early Warning (n=143), Moderate Risk (n=241), and Lower Risk (n=344) based on structural burden and overdose mortality.}\label{fig:silent}
\end{figure}

This produced four groups: High Risk/Crisis (n = 247), which includes counties with both high social disadvantage and high overdose deaths; Silent Risk/Early Warning (n = 143), which have high structural disadvantage but not yet high overdose mortality; Moderate Risk (n = 241), which have lower social disadvantage but still show unexpectedly high overdose deaths; and Lower Risk (n = 344; Figure~\ref{fig:silent}), which have both low social burden and low mortality.

When comparing Silent Risk counties with Lower Risk counties, we found significant differences across all 20 SDoH indicators (all $p < 0.001$), meaning the differences are statistically strong and not due to chance. Silent Risk counties showed higher levels of disadvantage in almost every factor, including higher minority population (41.9\% vs 29.1\%), higher poverty rates (27.6\% vs 16.5\%), more mobile homes (11.9\% vs 4.9\%), and much higher overall social vulnerability (SVI: 0.81 vs 0.38). They also had higher physical inactivity rates (27.8\% vs 21.0\%).

One key finding is that Silent Risk counties actually have fewer psychiatrists per 100,000 people (6.53 vs 10.73). This suggests weaker access to mental healthcare and overall treatment systems. It indicates that these counties are not only socially disadvantaged, but also lack enough healthcare infrastructure, which may limit their ability to respond when overdose risk increases. A ranked list of the top 25 counties with the largest gap between SDoH burden and mortality is shown in Table~\ref{tab:silent}.

\begin{sidewaystable}
\caption{Top 25 Silent Risk Counties Ranked by SDoH Burden Score}\label{tab:silent}
\begin{tabular}{@{}lllllllll@{}}
\toprule
County & SMR & Poverty (\%) & Uninsured (\%) & Depression (\%) & Smoking (\%) & Trt. Desert & Rural & SDoH Burden \\
\midrule
Liberty County, TX & 0.410 & 30.1 & 24.5 & 23.3 & 22.5 & No & No & 0.709 \\
Putnam County, FL & 0.895 & 39.0 & 16.6 & 21.1 & 27.1 & Yes & Yes & 0.704 \\
Cameron County, TX & 0.167 & 40.0 & 27.5 & 19.6 & 17.1 & No & No & 0.702 \\
Potter County, TX & 0.362 & 35.4 & 21.5 & 22.2 & 21.9 & No & No & 0.698 \\
Navajo County, AZ & 1.039 & 38.8 & 14.5 & 24.0 & 23.6 & No & Yes & 0.691 \\
Webb County, TX & 0.500 & 34.0 & 28.4 & 19.7 & 16.7 & No & No & 0.679 \\
St. Landry Parish, LA & 0.563 & 40.3 & 8.8 & 25.8 & 25.1 & No & Yes & 0.673 \\
Pike County, MS & 1.055 & 40.8 & 12.7 & 22.6 & 23.2 & Yes & Yes & 0.669 \\
Okanogan County, WA & 0.970 & 30.3 & 12.8 & 26.7 & 20.3 & Yes & Yes & 0.635 \\
Jefferson County, TX & 0.400 & 29.2 & 20.8 & 20.8 & 20.2 & No & No & 0.634 \\
Chesterfield County, SC & 1.053 & 32.8 & 12.4 & 23.5 & 23.3 & No & Yes & 0.633 \\
Sampson County, NC & 0.990 & 31.9 & 15.8 & 22.4 & 21.7 & No & Yes & 0.631 \\
San Juan County, NM & 0.729 & 37.9 & 12.6 & 20.6 & 22.6 & No & No & 0.627 \\
Duplin County, NC & 0.854 & 31.6 & 14.8 & 22.6 & 21.3 & No & Yes & 0.615 \\
Laurens County, GA & 1.011 & 35.3 & 13.5 & 20.9 & 22.7 & No & Yes & 0.612 \\
Danville city, VA & 0.990 & 38.7 & 9.1 & 20.4 & 23.4 & No & Yes & 0.611 \\
Bibb County, GA & 0.669 & 34.6 & 13.9 & 20.7 & 20.4 & No & No & 0.606 \\
White County, AR & 0.592 & 29.6 & 9.6 & 26.9 & 21.8 & No & Yes & 0.601 \\
El Paso County, TX & 0.390 & 31.8 & 21.8 & 19.3 & 14.8 & No & No & 0.601 \\
Wyandotte County, KS & 0.959 & 28.8 & 17.5 & 18.5 & 23.2 & No & No & 0.599 \\
Pottawatomie County, OK & 0.682 & 25.7 & 13.3 & 26.7 & 23.1 & Yes & Yes & 0.599 \\
Jasper County, MO & 0.539 & 29.0 & 13.8 & 25.3 & 22.4 & No & No & 0.597 \\
Yakima County, WA & 0.976 & 29.0 & 13.3 & 24.7 & 17.5 & No & No & 0.595 \\
Caddo Parish, LA & 0.256 & 33.4 & 7.5 & 24.5 & 21.8 & No & No & 0.595 \\
Nueces County, TX & 0.476 & 26.4 & 18.1 & 22.9 & 17.4 & No & No & 0.594 \\
\bottomrule
\end{tabular}
\footnotetext{\textit{Note:} SMR: Standardized Mortality Ratio; SDoH: Social Determinants of Health. Trt. Desert = county with zero psychiatrists per AHRF. Rural status based on HRSA classification. Counties ranked by SDoH burden score in descending order.}
\end{sidewaystable}

\subsection{Spatial Autocorrelation}\label{subsec3f}

Global spatial autocorrelation analysis confirmed strong positive clustering, meaning counties with similar overdose mortality rates tended to be located near each other (Moran's I = 0.5053, z = 19.58, p=0.001). This means high-overdose counties were often located near other high-overdose counties, while low-overdose counties were often near other low-overdose counties. These results show that the clustering pattern was very strong and highly unlikely to have occurred by chance.

\begin{figure}[H]
\centering
\includegraphics[width=\textwidth]{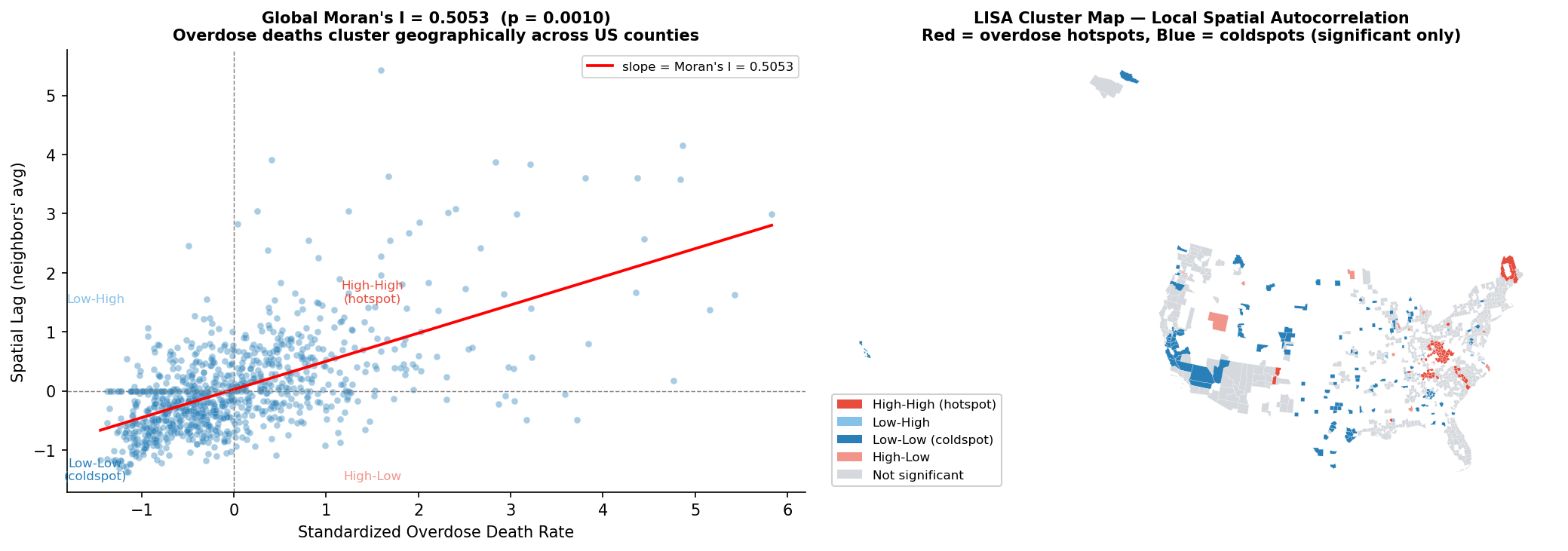}
\caption{LISA cluster map showing spatial autocorrelation of county-level fentanyl overdose mortality. High-High hotspot clusters (n=75) are concentrated in the Appalachian corridor, while Low-Low coldspot clusters (n=136) are concentrated in the Great Plains and Upper Midwest. Map lines delineate study areas and do not necessarily depict accepted national boundaries.}\label{fig:spatial}
\end{figure}

Local Indicators of Spatial Association identified the specific locations of these clusters. A total of 75 hotspot counties (High-High clusters), meaning high-overdose counties surrounded by other high-overdose counties, were mainly concentrated in the Appalachian corridor, including West Virginia, eastern Kentucky, southern Ohio, and eastern Tennessee. In contrast, 136 coldspot counties (Low-Low clusters), meaning low-overdose counties surrounded by other low-overdose counties, were primarily concentrated in the Great Plains and Upper Midwest, including Nebraska, Iowa, Kansas, and Minnesota. (Figure~\ref{fig:spatial}; map lines delineate study areas and do not necessarily depict accepted national boundaries).

\subsection{Suppressed County Analysis}\label{subsec3g}

CDC WONDER suppresses county overdose mortality data when there are fewer than 10 overdose deaths in a county, mainly to protect privacy. In the 2022 dataset, 2,351 U.S. counties had mortality records, but 1,368 counties (58.2\%) had suppressed data (Table~\ref{tab:suppressed}). This means more than half of U.S. counties could not be fully analyzed because their overdose counts were hidden.

Suppressed counties were much smaller, with an average population of 30,810 compared to 287,288 in non-suppressed counties. They were also far more rural, with 71.5\% classified as rural compared to 26.4\% of non-suppressed counties (Table~\ref{tab:suppressed}). This suggests that data suppression mainly affects rural America, which may cause overdose research to focus more on larger or urban counties where data is available.

\begin{table}[h]
\caption{Comparison of Observed and Suppressed Counties on Key SDoH Indicators}\label{tab:suppressed}
\begin{tabular}{@{}lllll@{}}
\toprule
Variable & Observed (n=983) & Suppressed (n=1368) & p-value & Sig \\
\midrule
Poverty Rate (\%) & $21.90 \pm 7.14$ & $25.12 \pm 8.37$ & 0.0000 & *** \\
Uninsured Rate (\%) & $8.16 \pm 3.68$ & $9.93 \pm 5.05$ & 0.0000 & *** \\
No HS Diploma (\%) & $10.49 \pm 4.44$ & $12.42 \pm 5.66$ & 0.0000 & *** \\
Minority Pop. (\%) & $29.15 \pm 19.65$ & $23.68 \pm 19.18$ & 0.0000 & *** \\
SVI Percentile & $0.54 \pm 0.27$ & $0.52 \pm 0.29$ & 0.4099 & ns \\
Population (mean) & $287,288 \pm 551,791$ & $30,810 \pm 36,144$ & 0.0000 & *** \\
Rural County, n (\%) & 260 (26.4\%) & 977 (71.5\%) & 0.0000 & *** \\
Treatment Desert, n (\%) & 127 (12.9\%) & 887 (64.9\%) & 0.0000 & *** \\
\bottomrule
\end{tabular}
\end{table}
\noindent\footnotesize\textit{Note:} SVI: Social Vulnerability Index. Treatment Desert = zero psychiatrists per AHRF. Significance: *** $p < 0.001$, ns = not significant.
\normalsize

\begin{figure}[H]
\centering
\includegraphics[width=\textwidth]{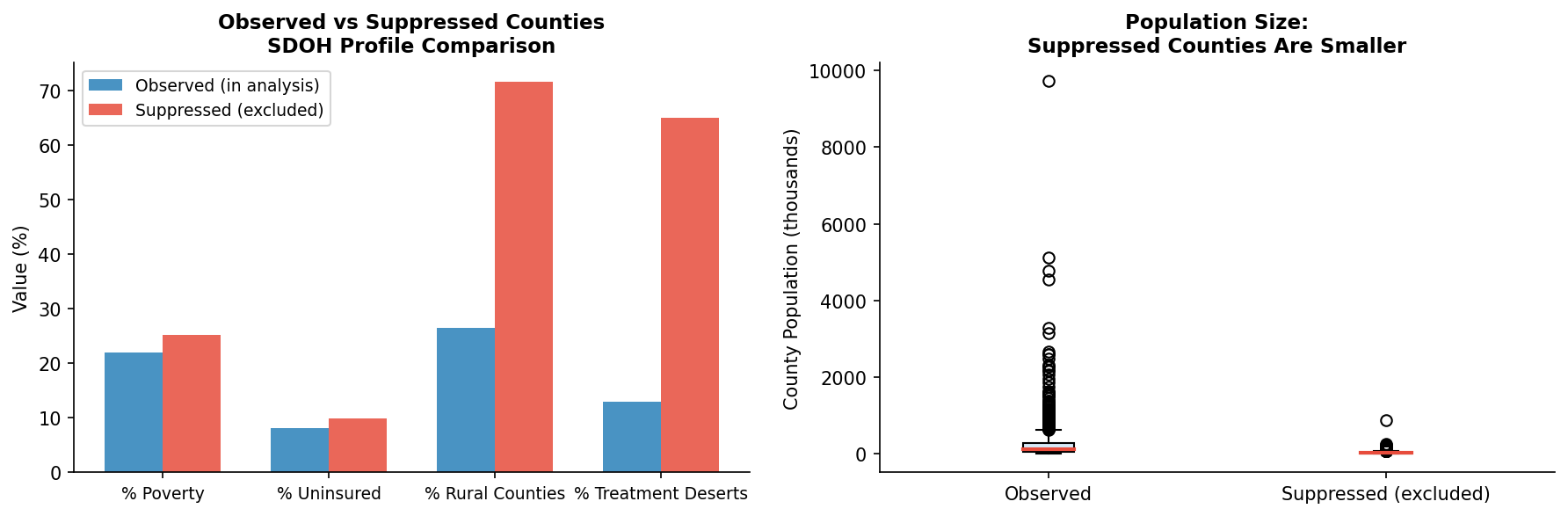}
\caption{Comparison of key SDoH indicators between observed and suppressed counties. Suppressed counties showed significantly higher poverty rates, lower educational attainment, and greater treatment desert prevalence compared to observed counties.}\label{fig:suppressed}
\end{figure}

Treatment desert status, defined as having no psychiatrist access, was much more common in suppressed counties, with 64.9\% lacking psychiatrists compared to 12.9\% of non-suppressed counties. Although overdose counts were hidden, these counties still showed high structural vulnerability. Suppressed counties also had higher poverty rates (25.1\% vs. 21.9\%) and lower educational attainment, with 12.4\% lacking a high school diploma compared to 10.5\% in non-suppressed counties (Figure~\ref{fig:suppressed}).

Notably, suppressed counties had lower minority population proportions (23.7\% vs. 29.2\%; Table~\ref{tab:suppressed}), showing that these hidden counties were more often small, rural, poor, and predominantly non-minority communities. These findings suggest that national overdose research may underestimate risk in small rural counties and low-resource geographic areas. Because many high-poverty rural counties are excluded due to suppression, national models may miss early-stage overdose risk, silent risk counties, and important structural vulnerability in rural America.

\section{Discussion}\label{sec4}

\subsection{Structural Vulnerability as the Dominant Predictive Signal}\label{subsec4a}
Structural health vulnerability findings showed that internal county characteristics, such as disability, hypertension, smoking, and chronic disease, were more important for predicting fentanyl overdose mortality in 2022 than simply where a county was located. Earlier research suggested that overdose spread was mainly driven by geographic diffusion \cite{bib10}. However, this shift is consistent with the ongoing fourth wave of the opioid epidemic \cite{bib5}. As fentanyl has spread across the entire country, location has become less important because fentanyl is now widespread. Instead, counties suffering the most are those with poverty, disability, smoking, weak healthcare access, and chronic disease. Mortality has become increasingly concentrated in counties with entrenched structural vulnerabilities rather than just proximity to earlier outbreak epicenters.

Disability as the strongest SHAP predictor was not a completely new finding. It replicated results from sub-city analyses of fentanyl mortality \cite{bib13} and extended this relationship to the national county level. Disability and chronic disease burden may act as broader markers of social and health disadvantage, representing multiple overlapping vulnerabilities such as reduced physiological resilience to overdose, limited healthcare engagement, weaker treatment access, and social isolation. These combined risks are often not fully captured by traditional regression models, which usually examine variables separately, while machine learning methods can better detect these interacting vulnerabilities.

The treatment desert mortality penalty (+0.616 SMR units; p<0.0001) directly contrasts with earlier findings \cite{bib10}, where buprenorphine provider waivers were considered the weakest predictive signal. This difference may be due to methodology. Earlier work modeled treatment access as a density gradient, while this study used a binary structural threshold based on whether counties had zero psychiatrists or not. Having no psychiatric providers may create a major structural disadvantage that functions more like a categorical shock than a gradual reduction in access. This type of nonlinear effect is something XGBoost is better suited to identify.

\subsection{Policy implications - treatment deserts and silent risk counties}\label{subsec4b}

Counties without psychiatric services, known as treatment deserts, had about 52\% higher overdose mortality than counties with psychiatrist access. Since about 54.9\% of U.S. counties lack psychiatric services, this treatment gap may contribute to thousands of potentially preventable deaths each year. Immediate expansion of OUD (opioid use disorder) treatment resources should prioritize the 127 treatment desert counties identified in this study, especially those with both high social and structural burden. Possible solutions include mobile treatment clinics, telehealth buprenorphine prescribing, and expanded primary care authority for addiction treatment. These counties are likely the highest priority for intervention because they have both poor treatment access and high vulnerability.

Prior research found that 71.2\% of rural counties lacked MOUD (medication for opioid use disorder) providers \cite{bib9}. The present study builds on this by showing that this shortage of treatment providers is also associated with higher overdose mortality.

The 143 silent risk counties identified in this analysis may be the best targets for early prevention because they already have strong structural disadvantages, but overdose deaths have not yet reached crisis levels. Table~\ref{tab:silent} provides a roadmap for preemptive public health action by identifying which counties need early support and what interventions may be most useful, including naloxone distribution, recovery support programs, and recruitment of additional addiction treatment providers before these counties enter active crisis.

Rosenblum and colleagues argued that the U.S. needs better systems to detect overdose threats earlier \cite{bib14}. The silent risk framework developed in this study offers a practical and low-cost way to support that goal by using existing public federal data.

\subsection{Spatial findings and implications}\label{subsec4c}

The strong spatial autocorrelation (Moran's I = 0.5053) shows that overdose deaths are not randomly distributed across the U.S. Instead, counties with high overdose mortality are often located near other high-mortality counties. This pattern is consistent with earlier research showing that overdose deaths follow geographic diffusion patterns over time \cite{bib15,bib16}, meaning the epidemic spreads in predictable regional clusters. The 75 high-risk counties are mainly concentrated in the Appalachian region, reflecting the established Appalachian--Midwest--New England epidemic core. These areas represent the main geographic center of the overdose crisis. A key methodological point is that many hotspot-adjacent counties are missing from the dataset because their data is suppressed in CDC WONDER. The suppressed county analysis shows that 72\% of these counties are rural and 65\% are treatment deserts. This suggests that many high-risk rural counties are not fully visible in the data.

Overall, this means the true geographic extent of the overdose epidemic is likely larger than what is shown in official data, because aggregate mortality data alone does not capture these hidden counties.

\subsection{Limitations}\label{subsec4d}

Firstly, the study uses a cross-sectional design with only 2022 data, so it cannot prove cause and effect. It can show that SDOH factors are associated with overdose mortality, but it cannot confirm that they directly cause it. Second, the analysis is done at the county level, not the individual level. This means we cannot assume individual behavior from county patterns. This is known as the ecological fallacy, for example, a county with high disability rates includes both healthy and disabled individuals, and overdose deaths may not occur mainly among people with disabilities. Third, the study uses publicly available CDC WONDER data, which led to the exclusion of 1,368 counties (58.2\%) due to data suppression. These missing counties are mostly rural and more likely to be treatment deserts. Because of this, the relationship between SDOH and mortality may be underestimated in the most vulnerable areas. Unlike \cite{bib10}, who used restricted CDC data to include all counties, this study prioritizes data accessibility and single-year analysis over full geographic coverage. Fourth, treatment desert status is defined only as zero psychiatrists in AHRF data. This does not include other forms of treatment access such as telehealth services, peer recovery programs, or primary care providers prescribing buprenorphine. Because of this, treatment availability in some rural areas may be underestimated. Fifth, while the XGBoost model performed well in ranking counties by risk (Spearman $\rho$=0.670, Pearson r=0.687, and high-risk recall=71.1\%), the exact prediction accuracy was moderate (R$^{2}$=0.457, MAE=0.409 SMR units, MAPE=42.6\%, and 43.3\% of predictions within $\pm$0.25 SMR). This reflects that predicting county-level mortality in a single year is difficult due to natural variation that cannot be fully explained by structural factors alone. For a median population county (124,237 residents), the MAE of 0.409 SMR is roughly equal to about 12 deaths above or below the predicted value per year. This does not reduce the usefulness of the model for identifying high-risk counties, but it does limit its ability to predict exact death counts. Studies reporting higher accuracy (\cite{bib11}: R$^{2}$=0.93) used multi-year data, which smooths out yearly variation and is not directly comparable.

\section{Conclusion}\label{sec5}

County-level fentanyl overdose death rates in U.S. can be predicted using publicly available social and health data (SDOH). Using XGBoost and SHAP attribution, this study found that several county conditions were especially important in predicting overdose mortality, including high disability rates, behavioral health burdens such as smoking and mental health challenges, and lack of treatment access, especially in counties with no psychiatric services. The 143 silent risk counties identified in this study already show serious structural disadvantages, even though overdose deaths have not yet reached crisis levels. These counties may face high future risk, so public health systems should act early by investing in treatment programs, naloxone distribution, and other prevention resources before mortality worsens. Future research will improve this framework through geographically weighted modeling and local SHAP analysis to better understand how overdose risk factors differ by location. Since not every county has the same drivers of overdose risk, future studies will examine which factors matter most in specific regions and how local conditions shape overdose patterns.

\backmatter

\bibliography{sn-bibliography}

\end{document}